\documentclass[10pt,twocolumn,letterpaper]{article}

\usepackage[pagenumbers]{cvpr} 

\usepackage{graphicx}
\usepackage{amsmath}
\usepackage{amssymb}
\usepackage{booktabs}

\usepackage{multirow}

\usepackage{xcolor}

\newcommand{\best}[1]{\textbf{#1}}

\definecolor{Gray}{gray}{0.95}
\usepackage{colortbl}
\newcolumntype{a}{>{\columncolor{Gray}}c}
\newcommand{\mc}[2]{\multicolumn{#1}{c}{#2}}
\usepackage{bm}

\DeclareMathOperator*{\argmin}{arg\,min}

\usepackage{pifont}   
\usepackage[pagebackref,breaklinks,colorlinks]{hyperref}

\usepackage[capitalize]{cleveref}
\crefname{section}{Sec.}{Secs.}
\Crefname{section}{Section}{Sections}
\Crefname{table}{Table}{Tables}
\crefname{table}{Tab.}{Tabs.}

\newcommand*{\affaddr}[1]{#1} 
\newcommand*{\affmark}[1][*]{\textsuperscript{#1}}
\newcommand*{\email}[1]{\texttt{#1}}
\usepackage[symbol]{footmisc}

\def\cvprPaperID{*****} 
\def\confName{xxxx}
\def\confYear{2023}

\begin{document}

\title{Make-Your-Video:\\ Customized Video Generation Using Textual and Structural Guidance}

\author{%
Jinbo Xing\affmark[1]~ 
Menghan Xia\affmark[2,]$^*$~
Yuxin Liu\affmark[1]~ 
Yuechen Zhang\affmark[1]~
Yong Zhang\affmark[2]~
Yingqing He\affmark[3]~
Hanyuan Liu\affmark[1]~\\
Haoxin Chen\affmark[2]~
Xiaodong Cun\affmark[2]~
Xintao Wang\affmark[2]~
Ying Shan\affmark[2]~
Tien-Tsin Wong\affmark[1]\\
\affaddr{\affmark[1]CUHK~~~~~~~~~~}
\affaddr{\affmark[2]Tencent AI Lab~~~~~~~~~~}
\affaddr{\affmark[3]HKUST}\\
\small\email{$^*$Corresponding Author. Project page: \url{https://doubiiu.github.io/projects/Make-Your-Video}}
}

\twocolumn[{
\maketitle

\begin{center}
    \captionsetup{type=figure}
    \includegraphics[width=1\textwidth]{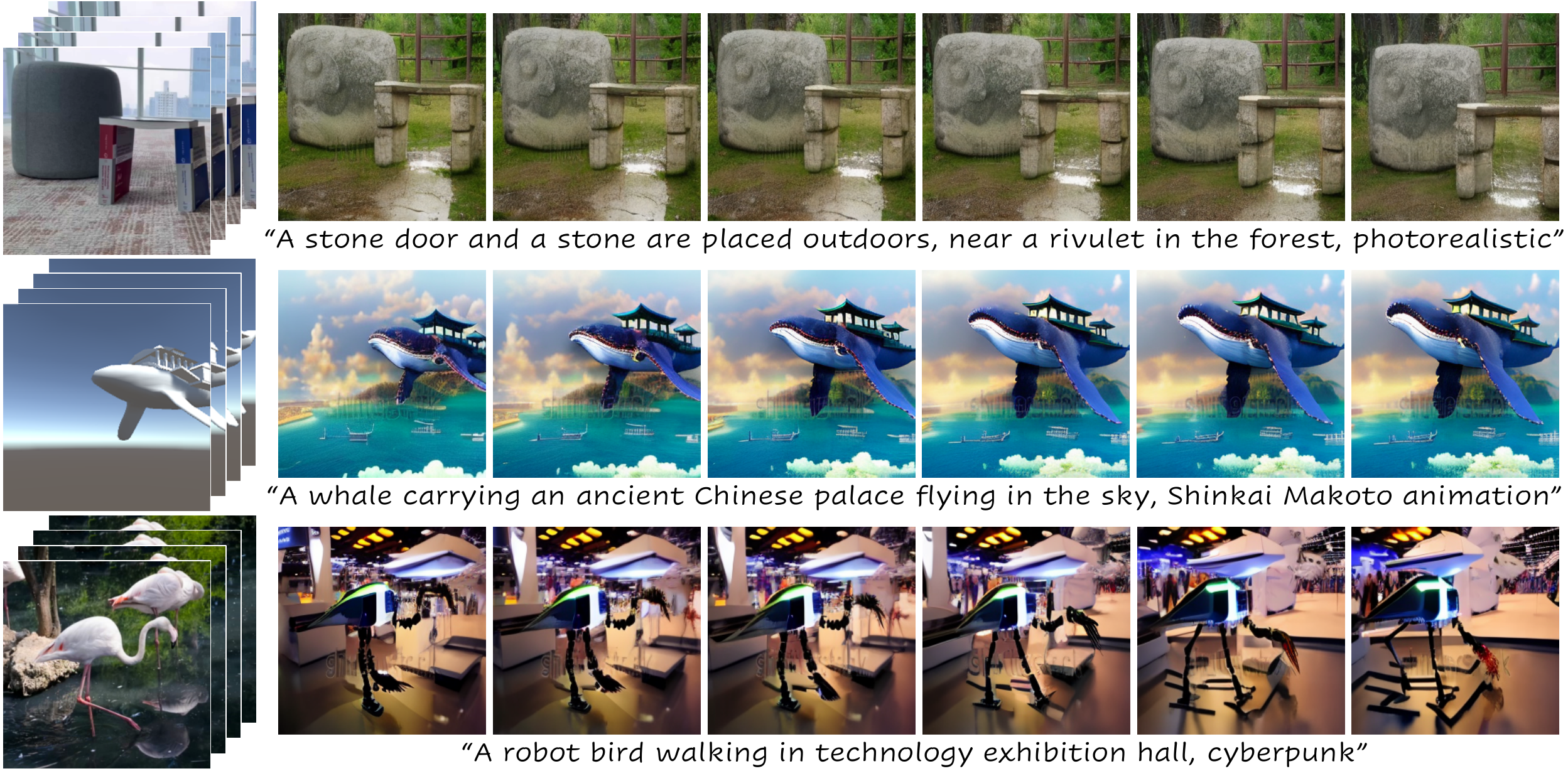}
    \captionof{figure}{Given the text descriptions and motion structure as guidance, our model can generate temporally coherent videos adhering to the guidance intentions. By building structural guidance from distinct sources, we show the video generation results in different applications, including (top) real-life scene setup to video, (middle) dynamic 3D scene modeling to video, and (bottom) video re-rendering.
    }
    \label{fig:teaser}
\end{center}
}]
\begin{abstract}
Creating a vivid video from the event or scenario in our imagination is a truly fascinating experience. Recent advancements in text-to-video synthesis have unveiled the potential to achieve this with prompts only. While text is convenient in conveying the overall scene context, it may be insufficient to control precisely. In this paper, we explore customized video generation by utilizing text as context description and motion structure (e.g. frame-wise depth) as concrete guidance.
Our method, dubbed Make-Your-Video, involves joint-conditional video generation using a Latent Diffusion Model that is pre-trained for still image synthesis and then promoted for video generation with the introduction of temporal modules. This two-stage learning scheme not only reduces the computing resources required, but also improves the performance by transferring the rich concepts available in image datasets solely into video generation. Moreover, we use a simple yet effective causal attention mask strategy to enable longer video synthesis, which mitigates the potential quality degradation effectively.
Experimental results show the superiority of our method over existing baselines, particularly in terms of temporal coherence and fidelity to users' guidance. In addition, our model enables several intriguing applications that demonstrate potential for practical usage. 
\end{abstract}

\section{Introduction}
\label{sec:intro}

As a widely embraced digital medium, videos are highly regarded for their ability to deliver vibrant and immersive visual experiences. Capturing real-world events in video has become effortless with the widespread availability of smartphones and digital cameras. However, when it comes to creating a video to express the idea aesthetically, the process becomes considerably more challenging and costly, which usually requires professional expertise in computer graphics, modeling, and animation production. Fortunately, recent advancements in text-to-video~\cite{ho2022imagen,singer2022make} shed light on the possibility of simplifying this process as textual prompts alone. Although text is recognized as a standard and versatile description tool, we argue that it excels primarily in conveying abstract global context, while it may be less effective in providing precise and detailed control. This motivates us to explore customized video generation by utilizing text as context description and motion structure as concrete guidance.

Specifically, we choose frame-wise depth maps to represent the motion structure, as they are 3D-aware 2D data that align effectively with the task of video generation. In our approach, the structural guidance can be quite rough, so as to allow non-professionals to easily prepare it. This design offers the flexibility for the generative model to produce plausible content without depending on intricately crafted input.
For instance, a scene setup using office common items can be used to guide the generation of a photorealistic outdoor landscape (Figure~\ref{fig:teaser}(top)). With 3D modeling software, the physical objects can be replaced with simple geometric elements or any accessible 3D assets (Figure~\ref{fig:teaser}(middle)). Naturally, another alternative is to make use of the estimated depth from existing videos (Figure~\ref{fig:teaser}(bottom)). So, the combination of textual and structural guidance provides users with both flexibility and controllability to customize their videos as intended.

To achieve this, we formulate the conditional video generation using a Latent Diffusion Model (LDM)~\cite{rombach2022high} that adopts a diffusion model in a compressed lower-dimensional latent space to reduce the computational expenses. For training an open-world video generation model, we propose to separate the training of spatial modules (for image synthesis) and temporal modules (for temporal coherence). This design is based on two primary considerations: (i) training the model components separately eases the computational resource requirements, which is especially crucial for resource-intensive tasks; (ii) as image datasets encompass a much enriched variety of concepts than the existing video datasets, hence, pre-training the model for image synthesis helps to inherit the diverse visual concepts and transfers them to video generation. A major challenge is to achieve temporal coherence. Specifically, using a pre-trained image LDM, we maintain them as the frozen spatial blocks and introduce the temporal blocks dedicated to learning inter-frame coherence over the video dataset.
Notably, we combine the temporal convolutions with the additional spatial convolutions, which improves the temporal stability by increasing the adaptability to the pre-trained modules. We also adopt a simple yet effective causal attention mask strategy to enable longer (i.e., 4$\times$ the training length) video synthesis and it mitigates the potential quality degradation significantly.

Both qualitative and quantitative evaluations evidence the superiority of our proposed method over existing baselines, particularly in terms of temporal coherence and fidelity to users' guidance. Ablation studies confirm the effectiveness of our proposed designs, which play a crucial role in the performance of our method. Additionally, we showcased several intriguing applications facilitated by our approach, and the results indicate the potential for practical scenarios.

Our contributions are summarized as below:
\begin{itemize}
    \item We present an efficient approach for customized video generation by introducing textual and structural guidance. Our method achieves top performance in controllable text-to-video generation both quantitatively and qualitatively.
    \item We propose a mechanism to leverage pre-trained image LDMs for video generation, which inherits the rich visual concepts while achieving a decent temporal coherence.
    \item We introduce a temporal masking mechanism to allow longer video synthesis while alleviating the quality degradation.
\end{itemize}
\section{Related Work}
\subsection{Diffusion Models for Text-to-Image (T2I) Synthesis}
Diffusion models~\cite{sohl2015deep,ho2020denoising,song2021score} (DMs) have recently shown unprecedented generative semantic and compositional power, attracting attention from both academia and industry. By absorbing the text embedding from CLIP~\cite{radford2021learning} or T5~\cite{raffel2020exploring}, they have been successfully adopted for text-to-image synthesis~\cite{ramesh2022hierarchical,nichol2022glide,saharia2022photorealistic,gu2022vector,rombach2022high,balaji2022ediffi}. GLIDE~\cite{nichol2022glide} introduces classifier-free guidance~\cite{ho2021classifier} to the text-conditioned image synthesis DMs, improving image quality in both photorealism and text-image alignment, which are further boosted by using CLIP~\cite{radford2021learning} feature space in DALL$\cdot$E 2. Moreover, a line of works~\cite{zeng2022scenecomposer,yang2022reco,li2023gligen,zhang2023adding,mou2023t2i} improves the controllability of T2I by introducing additional conditional inputs, \eg, pose, depth, and normal map. Since DMs generally require iterative denoising processes through a large U-Net, the training becomes computationally expensive. To address this, cascaded (Imagen~\cite{saharia2022photorealistic}) and latent diffusion models (LDMs~\cite{rombach2022high}) have been proposed. Specifically, Imagen adopts cascaded diffusion models in pixel space to generate high-definition videos, while LDMs first compress the image data using an autoencoder and learn the DMs on the resultant latent space to improve efficiency. To inherit the diverse visual concepts and reduce the training cost, our method builds upon LDMs by introducing video awareness through additional learnable temporal modules into the architecture and training on text-video data, while keeping the original weights of LDMs frozen.

\subsection{Diffusion Models for Text-to-Video (T2V) Synthesis}
Although there have been significant advancements in T2I, text-to-video generation (T2V) is still lagging behind due to the scarcity of large-scale high-quality paired text-video data, the inherent complexity of modeling temporal consistency, and resource-intensive training. As a pioneering work, Video Diffusion Model~\cite{ho2022video} models low-resolution videos with DMs using a space-time factorized U-Net in pixel space and trains jointly on image and video data. To generate high-definition videos, Imagen-Video~\cite{ho2022imagen} proposes effective cascaded diffusion models and the $\mathbf{v}$-prediction parameterization method. 

To reduce the training cost, many subsequent studies, including Make-A-Video~\cite{singer2022make}, MagicVideo~\cite{zhou2022magicvideo} and LVDM~\cite{he2022latent}, transfer T2I knowledge to T2V generation by initiating from pre-trained T2I models and fine-tuning the entire model. Differently, Khachatryan~\etal~\cite{khachatryan2023text2video} present a training-free transfer by leveraging pre-trained T2I models with manual pseudo motion dynamics to generate short videos. However, the generated videos suffer from low quality and inconsistency. Besides adopting the pre-trained T2I, Gen-1~\cite{esser2023structure} and FollowYourPose~\cite{ma2023follow} propose to control the structure and motion dynamics of synthesized videos by depth and pose, respectively. With the same conditions as us, Gen-1, however, trains the entire model, which can be both time- and resource-consuming, potentially leading to a degradation of the inherited rich visual concepts.
As a concurrent work, Video-LDM~\cite{blattmann2023align} shares a similar motivation with ours, \ie, extending the image LDMs to video generators by introducing temporal layers and keeping the original weights frozen. However, solely using temporal layers may not be sufficient for adapting LDMs to video generators.

As for longer video synthesis, interpolation~\cite{zhou2022magicvideo,singer2022make,he2022latent,blattmann2023align} and prediction~\cite{blattmann2023align,gu2023seer} strategies are commonly adopted in existing diffusion-based T2V approaches. Unlike prediction, interpolation does not increase the physical time span of synthesized videos but only makes enhance their smoothness. However, the current prediction mechanism in video diffusion models is still limited in the curated domain, \eg, driving~\cite{blattmann2023align} and indoor scene~\cite{gu2023seer}. Our work aims to efficiently and effectively adapt a pre-trained T2I model to a joint text-structure-guided video generator, and investigate general video prediction mechanisms for longer video synthesis.

\subsection{Text-driven Video Editing}
In recent studies, DMs have demonstrated their efficacy in image editing tasks, as evidenced by several works~\cite{meng2021sdedit,hertz2022prompt,couairon2022diffedit,kawar2022imagic,tumanyan2022plug}. However, their application to video editing on individual frames can result in temporal inconsistency issues. To address this, Text2LIVE~\cite{bar2022text2live} combines Layered Neural Atlases~\cite{kasten2021layered} and the proposed text-driven image editing method, allowing texture-based video editing but struggling to accurately reflect the intended edits. To improve the video quality, recent diffusion-based video editing methods rely on either pre-trained large-scale video diffusion models~\cite{molad2023dreamix}, which are usually inaccessible and hard to reproduce due to the unaffordable training, or the inversion~\cite{mokady2022null,song2020denoising} followed by attention manipulation mechanism~\cite{wu2022tune,qi2023fatezero,liu2023video} using pre-trained T2I model, rendering tricky prompt engineering or manual hyper-parameter tuning process. Although both our conditional video generator and video editing methods can edit video content, a notable distinction is their reliance on the original video (\eg, for inversion purposes), whereas our model does not necessitate the source video as input.
\section{Method}
\label{sec:method}

\begin{figure*}[t]
    \centering
    \includegraphics[width=0.95\linewidth]{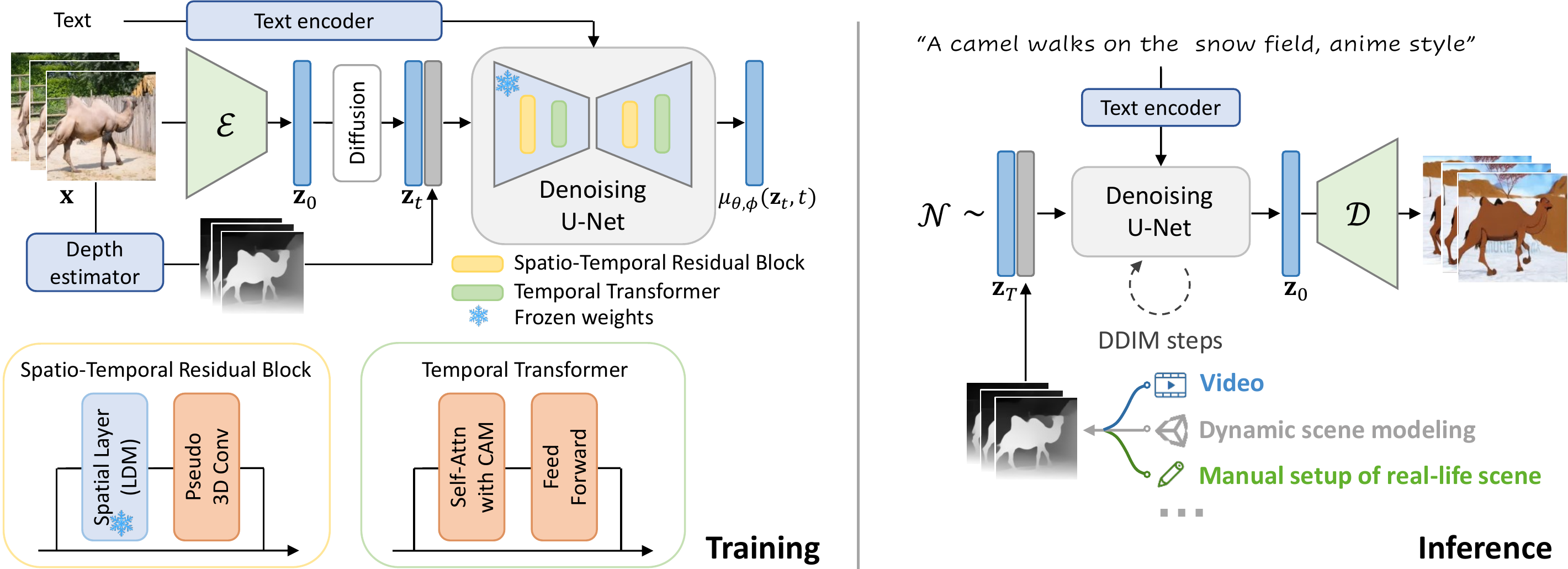}
    \caption{Flowchart of the proposed method. During training (left), the input video $\mathbf{x}$ is first encoded into latent feature $\mathbf{z}_0$ with a fixed pre-trained encoder $\mathcal{E}$ and diffused to $\mathbf{z}_t$. Meanwhile, the depth sequence will be extracted with the off-the-shelf depth estimator MiDas and concatenated with $\mathbf{z}_t$, and the text is encoded by a frozen OpenCLIP text encoder. Then the model learns to reverse the diffusion process conditioned on the depth and text prompt. As for inference (right), videos can be generated by recurrently denoising a random tensor sampled from normal distribution, under the guidance of text prompt and frame-wise depth obtained in multiple ways.}
    \label{fig:overview}
\end{figure*}

The goal of this work is to study controllable text-to-video synthesis so that the generated video could align with the users' intention faithfully. To achieve this, we propose a conditional video generation model that takes text prompts and frame-wise depths as conditional input. The text prompt describes the video appearance and depth sequence specifies the overall motion structure.

In the following, we start with a brief introduction to DMs and LDMs (Section~\ref{subsec:dms}), as the preliminary knowledge to our proposed controllable text-to-video generation model (Section~\ref{subsec:video}). At last, we discuss on a temporal masking mechanism to facilitate longer video generation (Section~\ref{subsec:prediction}).

\subsection{Preliminaries}
\label{subsec:dms}
\textbf{Diffusion models (DMs)} are probabilistic models designed to learn a target data distribution $p_{\text{data}}(\mathbf{x})$ by gradually denoising a normally distributed variable. This denoising process corresponds to learning the reverse process of a fixed Markov Chain of length $T$ with denoising score matching~\cite{song2021score,hyvarinen2005estimation,lyu2012interpretation}. The most successful models in the image synthesis field rely on a reweighted variant of the variational lower bound on $p_{\text{data}}(\mathbf{x})$. These models can be interpreted as an equally weighted sequence of denoising autoencoders $\mathbf{\epsilon}_{\theta}(\mathbf{x}_t,t)$, parameterized with learnable parameters $\theta$; $t=1 \ldots T$, which are trained to predict a denoised variant of their input $\mathbf{x}_t$, where $\mathbf{x}_t$ is a noisy version of the input $\mathbf{x} \sim p_{\text{data}}$. The corresponding denoising objective is
\begin{equation}
    \mathbb{E}_{\mathbf{x} \sim p_{\text{data}}, \mathbf{\epsilon} \sim \mathcal{N}(\mathbf{0}, \bm{I}),t} \left[\Vert \mathbf{\epsilon} - \mathbf{\epsilon}_\theta(\mathbf{x}_t; \mathbf{c}, t) \Vert_2^2 \right],
    \label{eq:diffusionobjective}
\end{equation}
where $\mathbf{c}$ is optional conditioning information (e.g. text prompt), $t$ denotes a timestep uniformly sampled from $\{1,\ldots,T\}$, and $\mathbf{\epsilon}$ is the noise tensor used during the diffusion from $\mathbf{x}_0$ to $\mathbf{x}_t$. The neural backbone implementation of $\mathbf{\epsilon}_\theta(\circ; \mathbf{c}, t)$ is generally a 2D U-Net~\cite{ronneberger2015u} with cross-attention conditioning mechanisms.

\noindent\textbf{Latent diffusion models (LDMs)}~\cite{rombach2022high} are proposed to improve the computational and memory efficiency over a learned compact latent space instead of the original pixel space. It is realized through perceptual compression with an autoencoder $\mathcal{E}$ and $\mathcal{D}$ for efficient and spatially lower-dimensional feature representations. The autoencoder is defined to reconstruct inputs $\mathbf{x}$, such that $\hat{x}=\mathcal{D(\mathcal{E(\mathbf{x})})}\approx\mathbf{x}$. A DM can then be trained in the compressed latent space and turned into a LDM. The corresponding objective is similar to Eq.~\ref{eq:diffusionobjective}, except for replacing $\mathbf{x}$ with its latent representaton $\mathbf{z}=\mathcal{E}(\mathbf{x})$. 


\subsection{Adapting LDMs for Conditional T2V Generation}
\label{subsec:video}


\begin{figure}[t]
    \centering
    \includegraphics[width=1\linewidth]{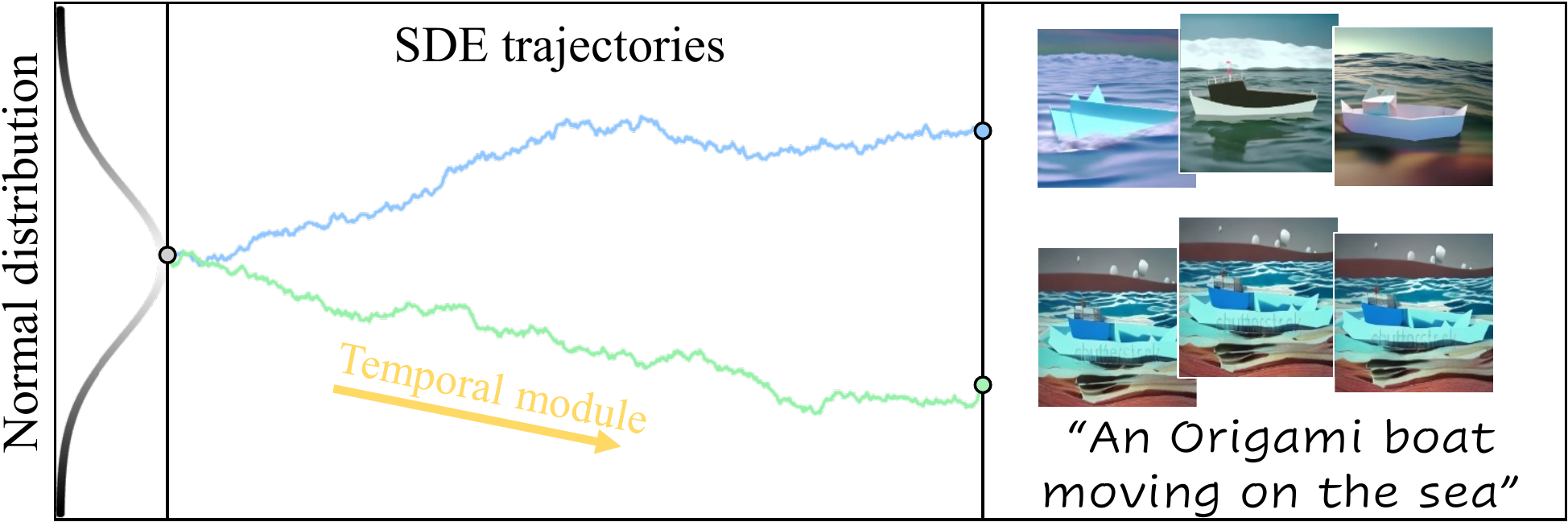}
    \caption{
    Concept depiction of the effect of temporal modules $\mathbf{\epsilon}_\phi$. Given a batch of noises, $\mathbf{\epsilon}_\phi$ pushes the SDE trajectories toward target data distributions with cross-frame coherence. We draw a single trajectory for a batch for conciseness.
    }
    \label{fig:sde}
\end{figure}


We employ an LDM to formulate our conditional video generation task, which involves synthesizing content for each frame while maintaining their temporal coherence to produce plausible dynamics.
Our key insight is to harness the power of pre-trained conditional T2I LDMs as a language-visual generative prior for conditional video synthesis. The main challenges are two-fold: (i) although the pre-trained image LDM can synthesize high-quality individual frames, it is not ready for generating temporally consistent video frames; and (ii) there is a shortage of large-scale text-video datasets with rich concept coverage and high quality when compared to image datasets like LAION~\cite{schuhmann2021laion}.
To address the first issue, we promote the original depth-conditioned LDM (CLDM) to video generators by introducing additional temporal modules. 
In principle, these modules push the reverse diffusion process of a pre-trained image CLDM toward a certain required subspace, which is conceptually shown in Figure~\ref{fig:sde}.
As for the second one, we freeze the weights of the pre-trained CLDM to lock the learned image prior and only fine-tune the temporal modules over a video dataset. The proposed framework is illustrated in Figure~\ref{fig:overview}.

Specifically, during training, the input video $\mathbf{x}$ is first encoded into the latent feature $\mathbf{z}_0 \in \mathbb{R}^{L\times C\times H^{\prime}\times W^{\prime}}$ in a frame-wise manner using pre-trained encoder $\mathcal{E}$, where $C$ is the number of latent channel dimensions, $H^{\prime}$ and $W^{\prime}$ are the latent spatial size, \ie, height and width, respectively. To inject structural guidance into denoising, we concatenate the frame-wise depth $\mathbf{s}$ extracted from the input video with an off-the-shelf depth estimator MiDas DPT-Hybrid~\cite{ranftl2020towards} with $\mathbf{z}_t$ that is diffused from latent feature $\mathbf{z}_0$.
The textural guidance is introduced through cross-attention layers in the OpenCLIP text embedding space.
The original image CLDM layers process the video data as a batch of independent input images by shifting the temporal dimension to the batch dimension for frame-wise processing. Accordingly, the batched feature output will be shifted back to video before being fed to the temporal modules.

We implement two types of temporal modules $\mathbf{\epsilon}_\phi$, namely, Spatio-Temporal Residual Block (STRB) and Temporal Transformer (TT), shown at the bottom of Figure~\ref{fig:overview}. STRB is an extension of the original residual block containing spatial layers only. To make it capable of capturing temporal priors in videos, we introduce additional pseudo 3D convolution layers, \ie, 2D spatial conv followed by 1D temporal conv, based on two considerations: (i) there is a domain gap between the modeled distribution from text-image dataset and target distribution in text-video data, as least in content; and (ii) the additional spatial layers could assist the temporal layers in learning motion dynamics by increasing the adaptability to those pre-trained spatial modules, which is evidenced by the experiments in Section~\ref{subsec:ablation}. The Temporal Transformer is located behind the original spatial transformer and its design is to leverage the property of temporal self-similarity to learn the inherent motion dynamics along the temporal axis in video data. Concretely, it consists of temporal self-attention modules with learnable positional embeddings~\cite{shaw2018self}, and a feed-forward layer~\cite{vaswani2017attention}.

Then our controllable T2V framework with denoiser $\mathbf{\epsilon}_{\theta,\phi}$ can be trained with a similar setting to the underlying CLDM. It is worth noting that the original weights $\mathbf{\epsilon}_\theta$ are frozen and only the added temporal modules $\mathbf{\epsilon}_\phi$ are learned via the optimization objective:
\begin{equation}
    \argmin_{\phi} \mathbb{E}_{\mathcal{E}(\mathbf{x}), \mathbf{\epsilon} \sim \mathcal{N}(\mathbf{0}, \bm{I}),t} \left[\Vert \mathbf{\epsilon} - \mathbf{\epsilon}_{\theta,\phi}(\mathbf{z}_t; \mathbf{c}, \mathbf{s}, t) \Vert_2^2 \right],
    \label{eq:videodiffusionobjective}
\end{equation}
where $\mathbf{z}_t$ indicates the diffused latent $\mathbf{z}=\mathcal{E}(\mathbf{x})$ with noise level $t$.

During inference, as depicted in Figure~\ref{fig:overview} (right), the depth $\mathbf{s}$ obtained from multiple sources for structure control and the text prompt $\mathbf{c}$ describing the target appearance serve as conditions for the reverse diffusion process. This process starts with randomly sampled noise $\mathbf{z}_T$ and the denoised latent $\mathbf{z}_0$ is converted into a video in pixel space using the pre-trained decoder $\mathcal{D}$.

\begin{figure}[t]
    \centering
    \includegraphics[width=0.8\linewidth]{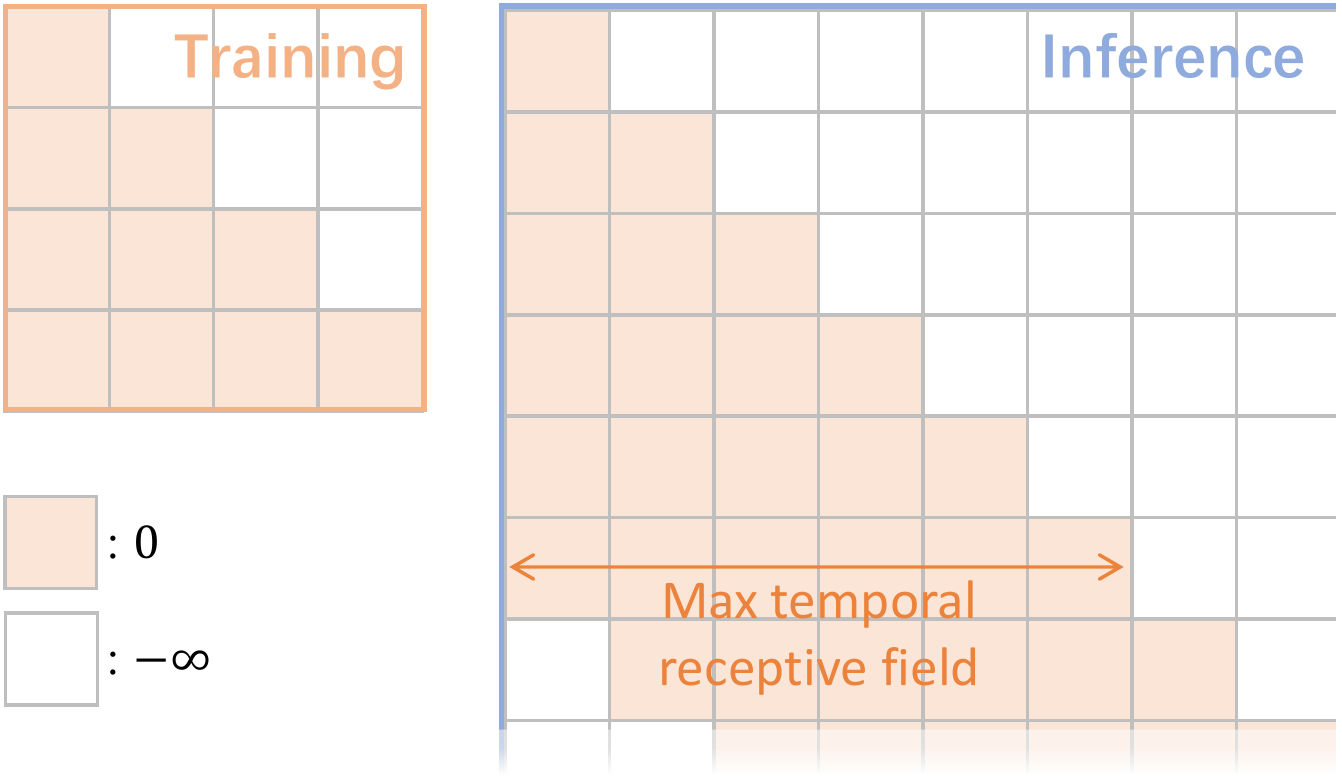}
    \caption{
    Illustration of the causal attention mask during training \& inference.
    }
    \label{fig:prediction}
\end{figure}

\subsection{Temporal Masking for Longer Video Synthesis}
\label{subsec:prediction}
Our conditional video LDM can generate satisfactory videos with the same number of frames (\ie, 16 frames) as in the training phase. However, when using it to generate longer videos during inference, we observe significant quality degradation. The possible reason is that the temporal self-attention is conducted in an N-to-N manner and the learned parameters are well-fitted to process a fixed length of tokens. According to this conjecture, longer token sequences tend to disturb each other because of the confused attention across frames.

To alleviate this problem, we propose to introduce a temporal masking mechanism so that the learned temporal attention module can better adapt to the cases involved in longer video synthesis. As shown in Figure~\ref{fig:prediction}, we adopt the causal attention mask (CAM) strategy to achieve this. 
The temporal attention $\mathbf{F}_t$ of an input feature $\mathbf{z}_t$ is calculated via:
\begin{equation}
    \mathbf{F}_t=\text{Attention}(\mathbf{Q}_t,\mathbf{K}_t,\mathbf{V}_t)=\text{softmax}(\frac{\mathbf{Q}_t\mathbf{K}_t^{\top}}{\sqrt{d}}+\mathbf{M})\mathbf{V}_t,
    \label{eq:cam}
\end{equation}
where $\mathbf{Q}_t$, $\mathbf{K}_t$, $\mathbf{V}_t$ are linearly projected features from $\mathbf{z}_t$, $d$ denotes the head dimension, and $\mathbf{M}$ is a lower triangular matrix ($\mathbf{M}_{i,j}=0$ if $i>j$ else $-\infty$) during training. For longer video synthesis during inference, the mask is modified to ensure the present token is only affected by the previous $L_M$ tokens, \ie, maximum temporal receptive field. With the help of CAM, the self-attention layers can be aware of different lengths of tokens, making the causal receptive field adjustable. It can thus effectively mitigate the quality degradation and temporal inconsistency problem for longer video synthesis. Still, larger $L_M$ leads to better cross-frame coherence while hinders the video quality, and vice versa.

Moreover, the CAM is an explicit way to inject video prior (\ie, motions in the video data are directional) into the adapted image LDM to learn temporal coherence, which could benefit even when generating short sequences, as evidenced in Section~\ref{subsec:ablation}. Thus we also adopt the causal attention mask in all experiments.

\section{Experiments}
\subsection{Implementation Details}
Our development is based on depth-conditioned Latent Diffusion Models~\cite{rombach2022high} (a.k.a Stable-Diffusion-Depth) implemented with PyTorch and the public pre-trained weights\footnote{\url{https://huggingface.co/stabilityai/stable-diffusion-2-depth}}. We train the newly introduced layers with 50K steps on the learning rate 1$\times$10$^{-4}$ and valid mini-batch size 512 with DeepSpeed~\cite{rasley2020deepspeed}. At inference, we use DDIM sampler~\cite{song2020denoising} with classifier-free guidance~\cite{ho2021classifier} in all experiments.

\textbf{Datasets.}
We use the WebVid-10M~\cite{bain2021frozen} dataset to turn the depth-conditioned LDM into a controllable text-to-video generator. WebVid-10M consists of 10.7 million video-caption pairs with a total of 52K video hours and is diverse and rich in content. During training, we sample 16 frames with a frame stride of 4 (assuming 30 FPS) and a resolution of 256 x 256 from input videos.

\begin{table}[t]
  \caption{Quantitative comparisons with state-of-the-art conditional video generation methods on UCF-101 for the zero-shot setting.}
  \label{tab:t2v_quan}
  \centering
  \begin{tabular}{lccc}
  \toprule
    Method  &Condition  & FVD $\downarrow$ & KVD $\downarrow$\\
    \midrule
    CogVideo (Chinese) &Text&  751.34& -\\
    CogVideo (English) &Text&  701.59& -\\
    MagicVideo &Text& 699.00& -\\
    Make-A-Video &Text&  367.23& -\\
    Video LDM &Text &  550.61& -\\
    \midrule
    T2V-zero+CtrlNet  &Text+Depth& 951.38 &115.55\\ 
    LVDM$_\text{Ext}$+Adapter &Text+Depth&  537.85& 85.47\\ %
    Ours w/o CAM & Text+Depth&390.63 & 36.57 \\
    \rowcolor{Gray}
    Ours &Text+Depth&  \best{330.49}&\best{29.52}\\
  \bottomrule
  \end{tabular}
\end{table}

\subsection{Evaluation on Video Generation}
Joint text-structure-conditioned video synthesis is a nascent area of computer vision and graphics, thus we find a limited number of publicly available research works to compare against. The extension version of LVDM~\cite{he2022latent} and T2V-zero~\cite{khachatryan2023text2video} are general text-to-video methods but capable of generating videos with additional conditions supported by ControlNet~\cite{zhang2023adding} or Adapter~\cite{mou2023t2i}, and we denote them as LVDM$_\text{Ext}$+Depth Adapter and T2V-zero+Depth CtrlNet, respectively. Meanwhile, we benchmark against pure text-to-video synthesis methods, including CogVideo~\cite{hong2023cogvideo}, MagicVideo~\cite{zhou2022magicvideo}, Make-A-Video~\cite{singer2022make}, and Video LDM~\cite{blattmann2023align}.
Since there is no structure control for these approaches, we include them here for reference and to examine the performance differences concerning structural guidance in text-to-video synthesis.

To evaluate the performance of video generation, we report the commonly-used Fr\'echet Video Distance (FVD)~\cite{unterthiner2019fvd} and Kernel Video Distance (KVD)~\cite{unterthiner2019fvd}, which evaluate video quality by measuring the feature-level similarity between synthesized and real videos based on the Fr\'echet distance and kernel methods, respectively. Specifically, they are computed by comparing 2K model samples (16 frames) with samples from the common evaluation~\cite{zhou2022magicvideo,blattmann2023align} dataset UCF-101~\cite{soomro2012ucf101}. Following~\cite{blattmann2023align}, we directly use UCF class names as text conditioning.

We evaluate in the zero-shot setting and tabulate the quantitative performance in Table~\ref{tab:t2v_quan}. According to the results, our method significantly outperforms all baselines with lower FVD and KVD. It is worth noting that although Make-A-Video with text condition only achieves an FVD value close to ours, they train on an additional extremely-large-scale dataset containing 100M text-video pairs, \ie, HD-VILA-100M~\cite{xue2022hdvila}. The superiority of our method indicates the effectiveness of the proposed video generation framework with textual and structural guidance and the adapting strategy that turns image LDMs into video generators. The qualitative comparison is made in the context of several application scenarios, as shown in Figure~\ref{fig:video_creation} and Figure~\ref{fig:video_rerendering}.

\begin{figure}[t]
    \centering
    
    \includegraphics[width=1\linewidth]{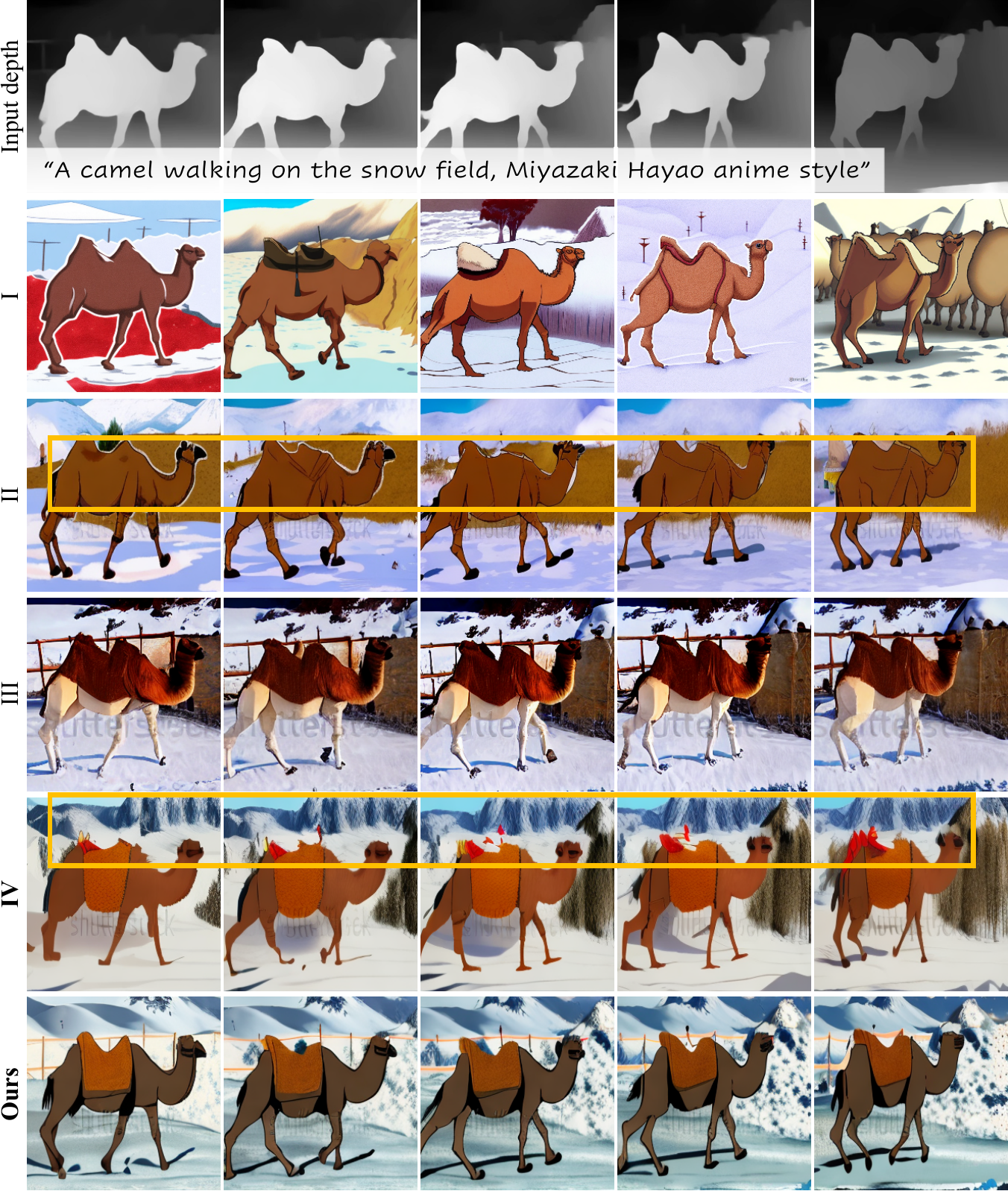}
    \caption{
    Visual comparisons of the videos synthesized by different variants of our approach.
    }
    \label{fig:ablation_adapting}
    
\end{figure}

\subsection{Ablation Studies}
\label{subsec:ablation}
We study several key designs of our proposed method in this section, including the adapting strategy and causal attention mask.

\begin{table}[t]
  \caption{Ablation study of different adapting strategies on UCF-101 for zero-shot setting. (Note: TT=Temporal Transformer, TC=1D Temporal Conv, P3D=Pseudo 3D Conv)}
  \label{tab:ablation_adapt}
\resizebox{\linewidth}{!}{
  \setlength{\tabcolsep}{2.1pt}
  \centering
  \begin{tabular}{lcccc}
  \toprule
    Variant  & Fine-tuned para. & FVD $\downarrow$ & KVD $\downarrow$ \\
    \midrule
   I. SD-Depth   & None     &1422.30 & 316.25\\
   II. \;\;~w/ TT     & TT    & 803.24 &  89.26\\
   III. \;\,w/ TT & Full U-Net  &  500.96 & 77.72\\ 
   IV. \;~w/ TT+TC  & TT+TC & 443.63  & 40.86\\
    \rowcolor{Gray}
   V. \;\;\:\!~w/ TT+P3D (Ours)& TT+P3D &\best{330.49}&\best{29.52}\\
  \bottomrule
  \end{tabular}
}
\end{table}

\textbf{Adapting strategy.} To study the effectiveness and superiority of our adapting strategy, we construct several baselines: \textbf{(I.)}: SD-Depth, the pre-trained image LDM that we build upon, \textbf{(II.)}: adding Temporal Transformer (TT) to baseline (I.) and fine-tuning this module, \textbf{(III.)}: the same architecture as (II.) but fine-tune the entire model, \textbf{(IV.)}: adding both TT and 1D Temporal Convolutions (TC) to (I.) and fine-tuning these two modules, and \textbf{(V.)}: introducing both TT and Pseudo 3D modules to (I.) and fine-tuning them, which is our full method. 
The quantitative comparison is shown in Table~\ref{tab:ablation_adapt}. By comparing baselines (I.), (II.), and (IV.), we can observe the improved performance in terms of both FVD and KVD by introducing more temporal modules that increase the temporal modeling capability.
The comparison between (V.) and (IV.) highlights the benefits of incorporating additional spatial layers with TC, which enhances the adaptability between the newly introduced temporal modules and the fixed spatial modules.
It is worth noting that although fine-tuning the entire model (III.) improves the performance quantitatively over its counterpart (II.), it cause a severe concept forgetting issue, as evidenced by Figure~\ref{fig:ablation_adapting} where (III.) fails to reflect the `anime style'. The visual comparison also tells the superiority of our full method in terms of both temporal coherence and conformity to the text prompt.

\begin{figure}[t]
    \centering
    \includegraphics[width=1\linewidth]{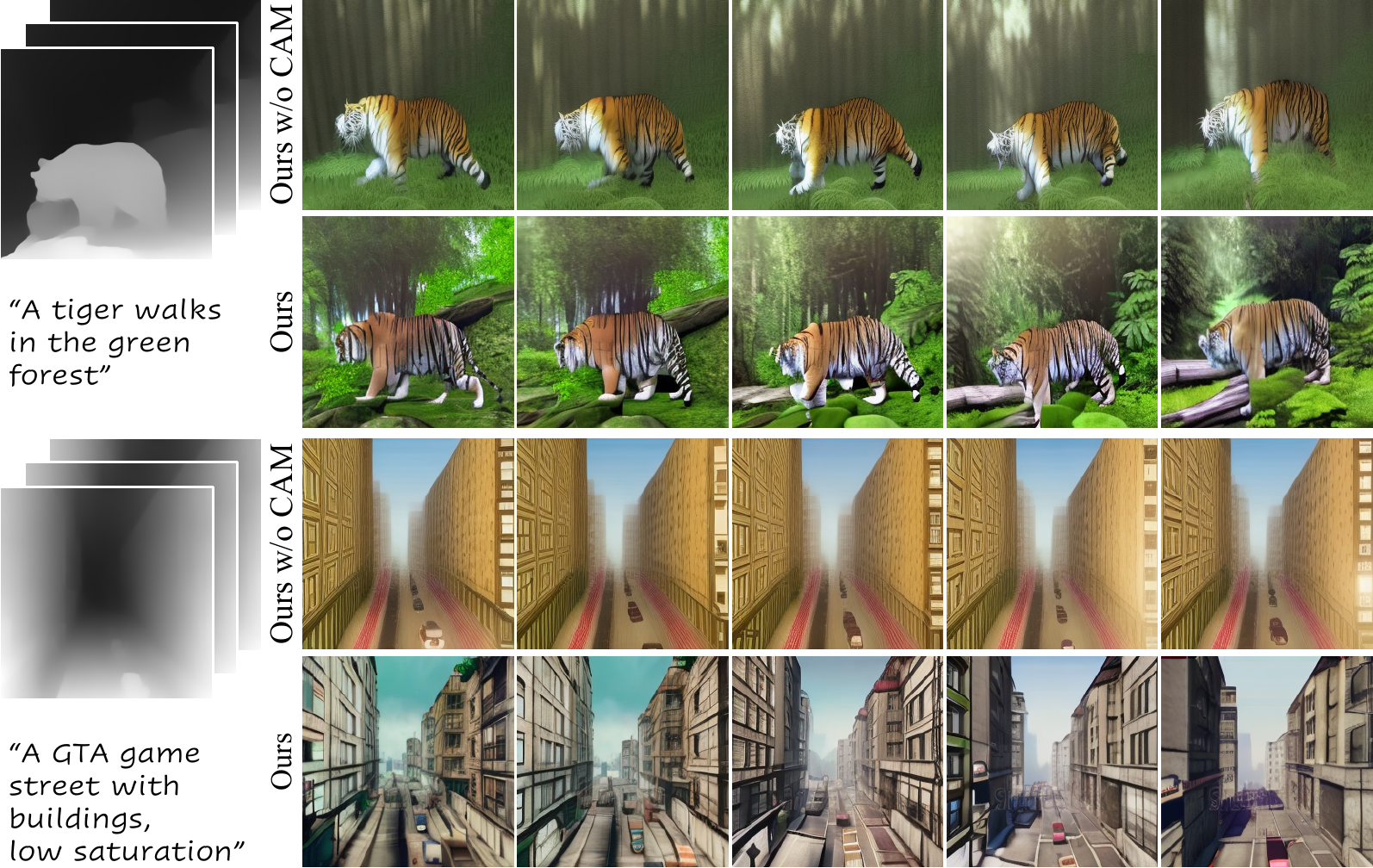}
    \caption{
    Visual comparisons of longer video synthesis (64 frames) produced by our baseline variant (w/o CAM) and our method (w/ CAM). Each frame is selected with a stride of 16.
    }
    \label{fig:ablation_prediction}
\end{figure}

\textbf{Causal attention mask.} We further investigate the effectiveness of causal attention mask (CAM). We construct a baseline without CAM and train it with the same sequence length as our full method (16 frames). As shown in Table~\ref{tab:t2v_quan} (bottom), CAM can boost the performance of conditional video generation in a 16-frame setting, owing to the introduced directional video motion prior.
We directly apply the trained models for longer video synthesis (64 frames). The qualitative comparison is presented in Figure~\ref{fig:ablation_prediction}, wherein our baseline variant suffers from significant quality degradation and temporal inconsistency (more evident in the supplementary video) due to the mismatched training and inference setting. In contrast, our method could produce plausible longer videos with greater detail and improved cross-frame coherence. Readers are recommended to check our supplementary video for better comparison.

\section{Applications}
\label{sec:application}

\subsection{Video Creation}
\begin{figure*}[th]
    \centering
    \includegraphics[width=0.95\linewidth]{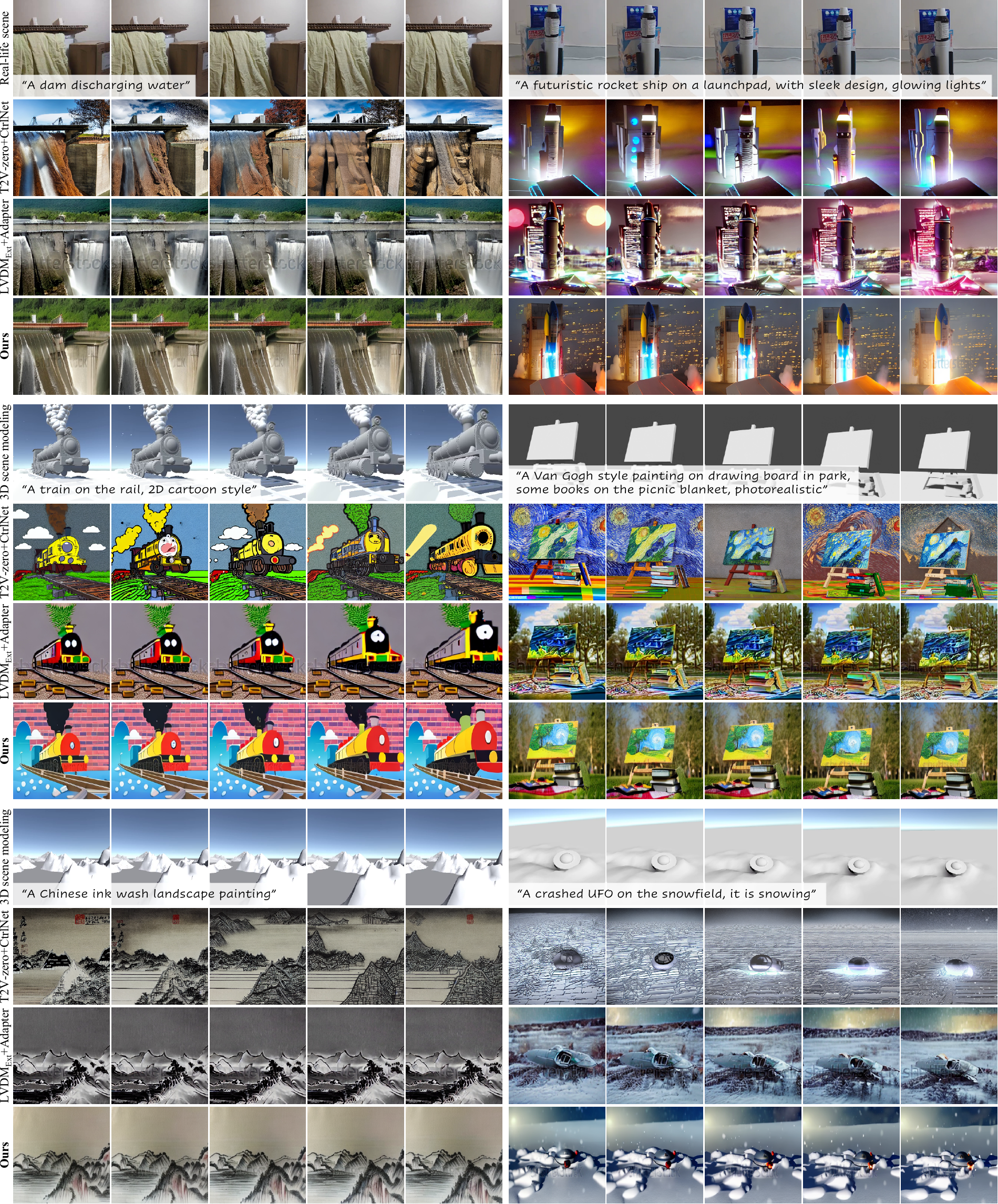}
    \caption{Visual comparison on videos generated in two applications, \ie, real-life scene to video (top) and 3D scene modeling to video (middle \& bottom).}
    \label{fig:video_creation}
    \vspace{5mm}
\end{figure*}

Our approach allows customized video creation guided by rough motion structures. So, it is feasible to capture a video of manually constructed miniature setup and use it as structural control for text-to-video synthesis. Some examples are illustrated in Figure~\ref{fig:teaser} (top) and Figure~\ref{fig:video_creation} (top). In comparison to other baseline methods, our approach is capable of generating high-fidelity, temporally consistent videos that closely adhere to the target textual descriptions and scene structure. In contrast, T2V-zero+CtrlNet mainly suffers from inconsistency due to its weak temporal constraint, and LVDM$_\text{Ext}$+Adapter tends to cause lower visual quality that inherits from the base text-to-video model. 
One can also construct the scene structure with 3D modeling software, \eg, Unity, as shown in Figure~\ref{fig:teaser} (middle) and Figure~\ref{fig:video_creation} (bottom). For LVDM$_\text{Ext}$+Adapter, apart from the aforementioned problems, it also fails to synthesize stylized videos, \eg, `2D cartoon' and `Chinese ink wash' at the bottom-left corner of Figure~\ref{fig:video_creation}. It is worth noting that our method can achieve both decent text alignment and cross-frame coherence.

\subsection{Video Re-rendering}
Video re-rendering here means changing the video appearance based on the text prompts while still preserving its motion structure. In addition to LVDM$_\text{Ext}$+Depth Adapter and T2V-zero+Depth CtrlNet, we also compare against the depth-conditioned image LDM (\ie, SD-Depth) that is used by our model as spatial modules, and a video editing method, Tune-A-Video~\cite{wu2022tune} combined with DDIM inversion~\cite{song2020denoising}, which requires per-video optimization and the original RGB video for inversion operation.
Following previous works~\cite{bar2022text2live,esser2023structure,wu2022tune}, we use videos from DAVIS~\cite{davis2017} and other in-the-wild videos, from which 11 representative videos with manually designed text prompts are utilized for evaluation.

\begin{figure*}[th]
    \centering
    \includegraphics[width=0.90\linewidth]{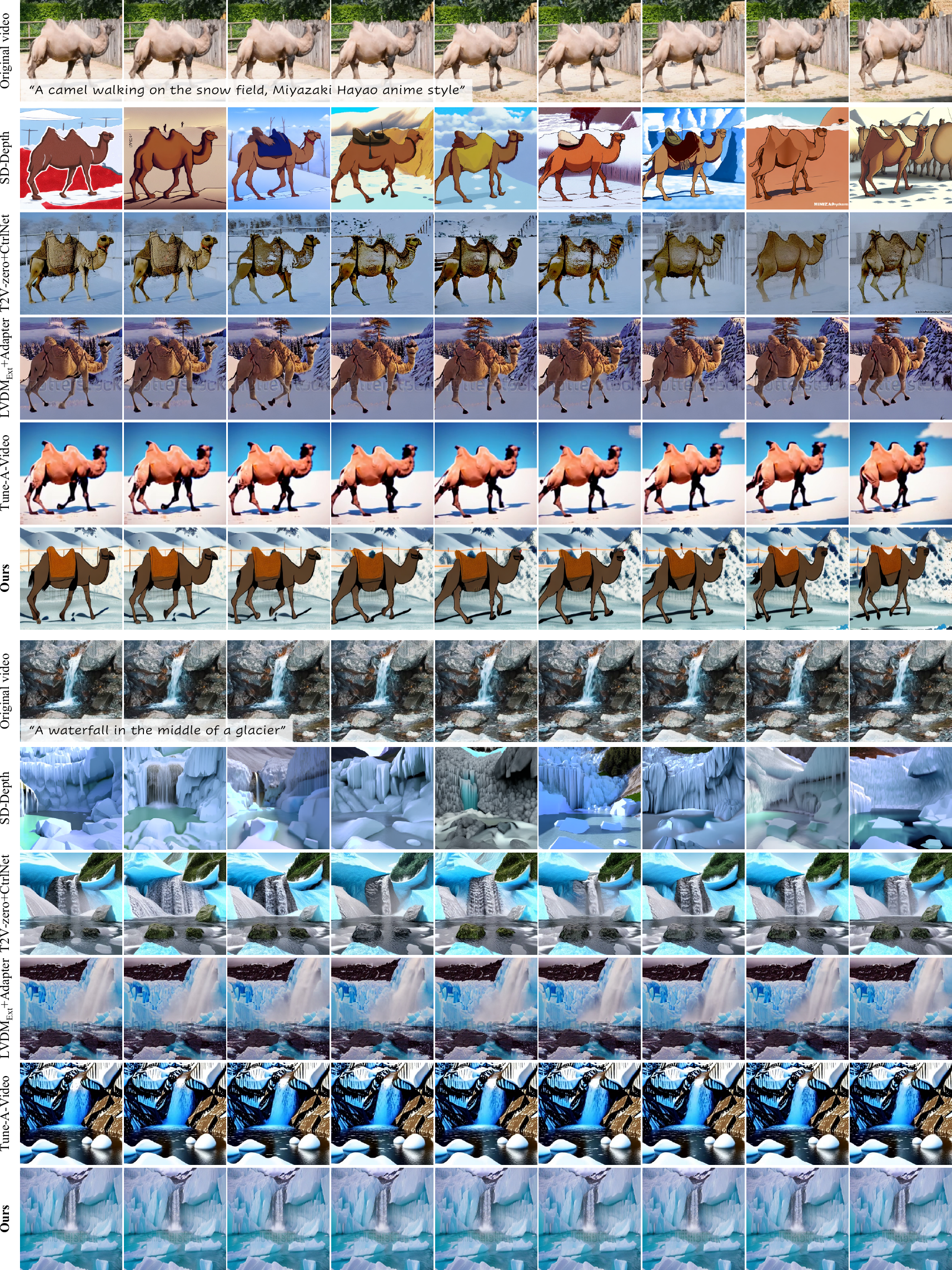}
    \caption{Visual comparison on examples of video re-rendering application, \ie, `camel' (top) and `waterfall' (bottom).}
    \label{fig:video_rerendering}
    \vspace{5mm}
\end{figure*}

\textbf{Quantitative evaluation.}
We measure temporal consistency and prompt conformity for performance comparison. Following~\cite{esser2023structure}, temporal consistency is calculated as the average cosine similarity between consecutive frame embeddings of CLIP image encoder, while prompt consistency is calculated as the average cosine similarity between text and image CLIP embeddings across all frames.
Table~\ref{tab:video_rerendering_quan} shows the results of each model, where `Temp.' and `Prompt' indicate the temporal coherence and prompt conformity, respectively. Our model outperforms the baseline models in temporal consistency and except SD-Depth in prompt conformity. Anyhow, as an image synthesis model, SD-Depth fails to synthesize consistent video frames. Besides, although achieves the second-best temporal consistency performance, Tune-A-Video suffers from overfitting to the original video, as evidenced by the results in Figure~\ref{fig:video_rerendering}.

\begin{table}[t]
  \caption{Quantitative comparisons for video re-rendering.}
  \label{tab:video_rerendering_quan}
\resizebox{\linewidth}{!}{
  \setlength{\tabcolsep}{2.1pt}
  \centering
  \begin{tabular}{lcccca}
  \toprule
    \multirow{2}{*}{Metric} & \multirow{2}{*}{SD-Depth} & {T2V-zero}  & LVDM$_\text{Ext}$&\multirow{2}{*}{Tune-A-Video} & \mc{1}{\multirow{2}{*}{Ours}} \\
    & &+CtrlNet &+Adapter && \mc{1}{}\\
    \midrule
    Temp.$\uparrow$ & 0.8624& 0.9109& 0.9613 &  0.9659&\best{0.9698}\\
    Prompt$\uparrow$ &\best{0.3695}& 0.3379  &0.3372&0.3473 &0.3547\\
  \bottomrule
  \end{tabular}
}
\end{table}

\textbf{Qualitative evaluation.}
We visually compare our method with other competitors in Figure~\ref{fig:video_rerendering}. In the `camel' case, the videos produced by our method are temporally coherent and well-align with the text description, \ie, `anime style'. In contrast, SD-Depth cannot produce consistent video frames; Tune-A-Video, LVDM$_\text{Ext}$+Adapter, and T2V-zero+CtrlNet suffer from low text-video conformity, visual concept forgetting, and structure deviations respectively. In `waterfall' case, our method shows similar superiority in the comprehensive quality including video quality, structure preservation, text-video conformity, and temporal coherence. Readers are recommended to check our supplementary video for better comparison.

\textbf{User study.} The human perception system is still the most reliable measure for video generation tasks. We conduct a user study, where 32 participants with good vision ability complete the evaluation successfully.
For each example, the participants are first shown with the input video and target prompt, followed by five randomly ordered videos re-rendered by different methods. Then, they are asked to rank the results in terms of temporal coherence, text \& structure guidance conformity, and frame quality: \{1: the best, 2: the second-best, $\ldots$, 5: the worst\}. We analyze the collected evaluation result in two aspects: (i) average ranking: the average ranking score of each method according to the rank-score table, and (ii) preference rate: the percentage of being selected as the best. The statistics are tabulated in Table~\ref{tab:user_study}. Our method earns the best ranking scores and preference rates in all three aspects.

\begin{table}[t]
  \caption{User study statistics of \emph{average ranking}$\downarrow$ and preference rate$\uparrow$.}
  \label{tab:user_study}
\resizebox{\linewidth}{!}{
  \setlength{\tabcolsep}{2.1pt}
  \centering
  \begin{tabular}{lcccca}
  \toprule
    \multirow{2}{*}{Property} & \multirow{2}{*}{SD-Depth} & {T2V-zero}  & LVDM$_\text{Ext}$&\multirow{2}{*}{Tune-A-Video} & \mc{1}{\multirow{2}{*}{Ours}} \\
    & &+CtrlNet &+Adapter && \mc{1}{}\\
    \midrule
    Temporal    & \emph{3.95}& \emph{3.40}& \emph{2.84} &  \emph{2.54}& \emph{\best{2.27}}\\
    coherence   & 6.53\%& 6.53\%&  12.78\%&  33.52\%& \best{40.62}\%\\
    \midrule
    Structure\& &  \emph{3.54}&  \emph{3.01}&   \emph{3.10}&  \emph{2.89} &  \emph{\best{2.45}}\\
    Text align.  & 13.07\%& 15.34\%& 12.50\% &  20.17\%& \best{38.92}\%\\
    \midrule
    Frame &  \emph{3.45}&  \emph{3.09}&  \emph{3.07} &   \emph{2.84}&  \emph{\best{2.55}}\\
    quality & 13.64\%& 13.07\%&  14.20\%& 24.15\% & \best{34.94}\%\\
  \bottomrule
  \end{tabular}
}
\end{table}
\section{Limitation}
Our method also has certain limitations. Firstly, our customized video generation model supports no precise control over visual appearance. For instance, it is intractable to synthesize videos featuring a specific individual or object. This relates to the concept customization technique, which has been extensively studied in T2I synthesis but remains less explored in T2V synthesis.
Additionally, our controllable text-to-video model explicitly demands frame-wise depth guidance, which can be overly intensive and may increase usage challenges in certain situations. A more efficient formulation could be supplying the model only with sparse keyframe guidance, which could help broaden the scope of potential applications. We leave these directions as future works.
\section{Conclusion}
We presented an efficient approach for customized video generation with textual and structural guidance. By employing a pre-trained image LDM as frozen spatial modules, our video generator exhibits a significant advantage in inheriting the wealth of visual concepts while maintaining satisfactory temporal coherence. Furthermore, we introduced a temporal masking mechanism to facilitate the synthesis of longer videos, which is a non-trivial extension for diffusion models. Ablation studies confirm the effectiveness of our proposed designs. Comparisons with state-of-the-art methods reveal our superiority in controllable text-to-video generation both quantitatively and qualitatively. Several applications show practical usage.

{\small
\bibliographystyle{ieee_fullname}
\bibliography{egbib}
}

\end{document}